\documentclass[a4paper,10pt,conference]{IEEEtran}

\usepackage{tablefootnote}
\usepackage{float}

\usepackage{cite}

\ifCLASSINFOpdf
   \usepackage[pdftex]{graphicx}
   \graphicspath{{figs/}}
   \DeclareGraphicsExtensions{.pdf,.jpeg,.png}
\else
   \usepackage[dvips]{graphicx}
   \graphicspath{{../figs/}}
   \DeclareGraphicsExtensions{.eps}
\fi
\usepackage{amsmath,amssymb,bm}
\usepackage{amsthm}
\usepackage{euscript}

\usepackage{booktabs}
\usepackage{bm}
\usepackage{color}
\usepackage{listings}
\usepackage{algorithmic}
\usepackage{array}

\usepackage{graphicx}
\usepackage{caption}
\usepackage{subcaption}

\usepackage{url}
\usepackage{multirow}
\usepackage[table,xcdraw]{xcolor}

\newif\iffinal
\finalfalse

\iffinal
\else
\usepackage[switch]{lineno}
\fi

\begin{document}
%
\title{Facial Point Graphs for Amyotrophic Lateral Sclerosis Identification}


\iffinal
\else
  \author{\IEEEauthorblockN{Nícolas Barbosa Gomes, Arissa Yoshida, Mateus Roder, Guilherme Camargo de Oliveira, \\Jo\~ao Paulo Papa}
		\IEEEauthorblockA{Departamento de Computa\c{c}\~ao\\
		Universidade Estadual Paulista (Unesp)\\
		Bauru, Brasil \\
		\{nicolas.gomes, arissa.yoshida, mateus.roder, gc.oliveira, joao.papa\}@unesp.br}	

  }
\fi

\maketitle

\begin{abstract}
    Identifying Amyotrophic Lateral Sclerosis (ALS) in its early stages is essential for establishing the beginning of treatment, enriching the outlook, and enhancing the overall well-being of those affected individuals. However, early diagnosis and detecting the disease's signs is not straightforward. A simpler and cheaper way arises by analyzing the patient's facial expressions through computational methods. When a patient with ALS engages in specific actions, e.g., opening their mouth, the movement of specific facial muscles differs from that observed in a healthy individual. This paper proposes Facial Point Graphs to learn information from the geometry of facial images to identify ALS automatically. The experimental outcomes in the Toronto Neuroface dataset show the proposed approach outperformed state-of-the-art results, fostering promising developments in the area.
\end{abstract}
\section{Introduction}
\label{s.introducao}

A gradual decline in the structure and functioning of the central nervous system marks Neurodegenerative Diseases (NDDs). The incidence and prevalence of these diseases exhibit a sharp increase with age, which means that life expectancy continues to rise in many parts of the world. Consequently, the number of cases is projected to grow in the future~\cite{checkoway2011neurodegenerative}. Despite the availability of certain treatments that can relieve the physical or mental symptoms linked to neurodegenerative diseases, there is currently no known method to slow down their progression or achieve a complete cure. 

Amyotrophic lateral sclerosis (ALS) is an NDD that causes the gradual deterioration of motor functions of the nervous system. Worldwide, the annual incidence of ALS is about 1.9 per 100,000 inhabitants~\cite{arthur2016projected}. In addition, patients face a delay in disease diagnosis by approximately 18 months~\cite{bandini2018automatic} and an average survival of 2 to 4 years after diagnosis~\cite{xu2021considerations}. Since effective treatments are currently unavailable, early and precise diagnosis is crucial in maintaining patients' quality of life.

Evaluating the facial expression of people is one effective way to diagnose neurodegenerative diseases, for the subject may lose a significant amount of verbal communication ability~\cite{yolcu2019facial}. It is worth noting that all types of NDDs affect the oro-facial musculature\footnote{Musculature related to communication and critical to functions such as chewing, swallowing, and breathing.} with significant impairments in speech, swallowing, and oro-motor skills, as well as emotion expression~\cite{bandini2020new}. Therefore, analyzing a patient's facial expression in an image or video can be crucial for diagnosing ALS.

The geometry-based characteristics derived from an individual's face describe the shape of its components, such as the eyes or mouth, which are very important for facial analysis~\cite{wu2019facial}. Based on these landmarks, Bandini et al.~\cite{bandini2018automatic} proposed an approach that predicts the patient's healthy state based on features representing motion, asymmetry, and face shape through video analysis. Such an inference was accomplished using well-known machine learning techniques, i.e., Support Vector Machines (SVM)~\cite{Cortes:95} and Logistic Regression~\cite{wright1995logistic}. Although reasonable results have been reported, there is still the need to deal with handcrafted features. Our work circumvents such a shortcoming by introducing Facial Point Graphs (FPGs) to learn motion information from facial expressions automatically. Our model is based on Graph Neural Networks (GNNs) and first constructs a graph with the most important facial points for ALS diagnosis to fulfill that purpose for further training. Later, each frame is classified as positive or negative to the disease. The majority voting then assigns the final label to individual.

As far as we know, no method employs Facial Point Graphs for ALS identification. We firmly believe that the landmarks extracted from frames can be better encoded in a non-Euclidean space, enabling the precise definition and representation of their distinct features. Therefore, the main contributions of this paper are twofold:
\begin{itemize}
    \item To introduce Facial Point Graphs to identify ALS.
    \item To employ a deep learning approach to the same context, thus not requiring handcrafted features.
\end{itemize}



The remainder of this paper is structured as follows: Sections~\ref{s.related_works} and~\ref{s.theoretical} present the literature review and theoretical background, respectively. Section~\ref{s.methodology} presents an explanation regarding the employed dataset, the used models to crop images and extract facial features, the proposed approach, and the classification method. Finally, Section~\ref{s.res} presents the experimental results and Section~\ref{s.discussion_conclusion} states the discussions about the results, conclusions, and future works.
\section{Related Works}
\label{s.related_works}

Facial expression is a significant part of human nonverbal contact, is more effective than words in face-to-face communication~\cite{mehrabian1968some}, and serves as a distinctive universal means of transmission. Very often, impaired facial expressions  manifest as symptomatic indications across countless medical conditions~\cite{yolcu2019facial}.

Bandini et al.~\cite{bandini2018automatic} introduced a novel approach for automatically detecting bulbar ALS. Their method involves analyzing facial movement features extracted from video recordings. The dataset comprises ten ALS patients (six male and four female) and eight age-matched healthy control subjects (six male and two female), which were asked to perform specific actions during recordings. Initially, each individual was recorded at rest (REST) with a neutral facial expression for 20 seconds. It is worth noting that this task was not used for analysis but only as a reference for extracting the geometric characteristics during the tasks.

Next, each participant was asked to perform the following actions: open their jaw to the maximum extent, repeated five times (OPEN); lip puckering (as if kissing a baby) a total of four times (KISS); pretend to blow out a candle, five times (BLOW); smile with closed lips, five times (SPREAD); repeat the syllable /pa/ in a single breath as fast as possible (PA); repeat the word /pataka/ as quickly as possible (PATAKA); repeat the sentence "Buy Bobby a puppy" (BBP) ten times in their usual tone and speaking speed.

Furthermore, the image pre-processing step was performed using the supervised descent method~\cite{xiong2013supervised}, which extracts corresponding facial landmarks for eyebrows, eyelids,  and nose, as likewise outer and inner lip contours for each frame. Also, a third coordinate was estimated for these landmarks based on intrinsic camera parameters. In this regard, feature extraction was carried out considering the points in the mouth region, as they demonstrated greater sensitivity to ALS. Considering aspects of lip movement such as range and speed of motion, symmetry, and shape, two different algorithms were used for classification: SVM and Logistic Regression. Last but not least, the best classification result was achieved in the BBP task, with an accuracy of 88.9\%.


Xu et al.~\cite{xu2020facial} conducted a study on classifying expressions using facial landmarks. Their approach used a Graph Convolutional Network (GCN)~\cite{kipf2016semi} to classify facial expressions in images. They employed the Dlib-ml machine learning algorithm~\cite{king2009dlib} to estimate the positions of 64 facial landmarks, which are employed to construct a graph along with their two-dimensional coordinates. The training process incorporated three different databases: JAFFE~\cite{lyons1998coding}, FER2013~\cite{goodfellow2013challenges}, and CK+~\cite{lucey2010extended}. The classes considered in this study included Anger, Disgust, Fear, Happiness, Sadness, and Surprise, achieving an accuracy of 95.85\%.


\section{Theoretical Background}
\label{s.theoretical}

\begin{figure}[ht]
  \subcaptionbox*{(a)}[.28\linewidth]{%
    \includegraphics[trim=0cm -5cm 0.8cm 0cm,clip, width=\linewidth]{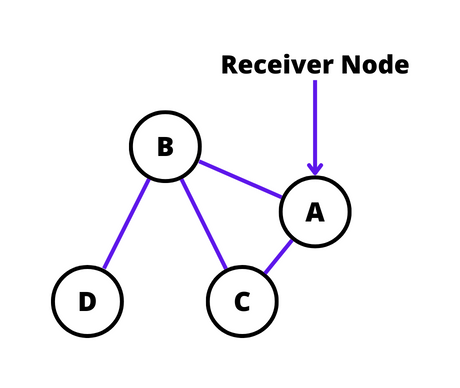}%
  }%
  \hfill
  \subcaptionbox*{(b)}[.77\linewidth]{%
    \includegraphics[trim=0cm 0cm 0cm 0.4cm,clip, width=\linewidth]{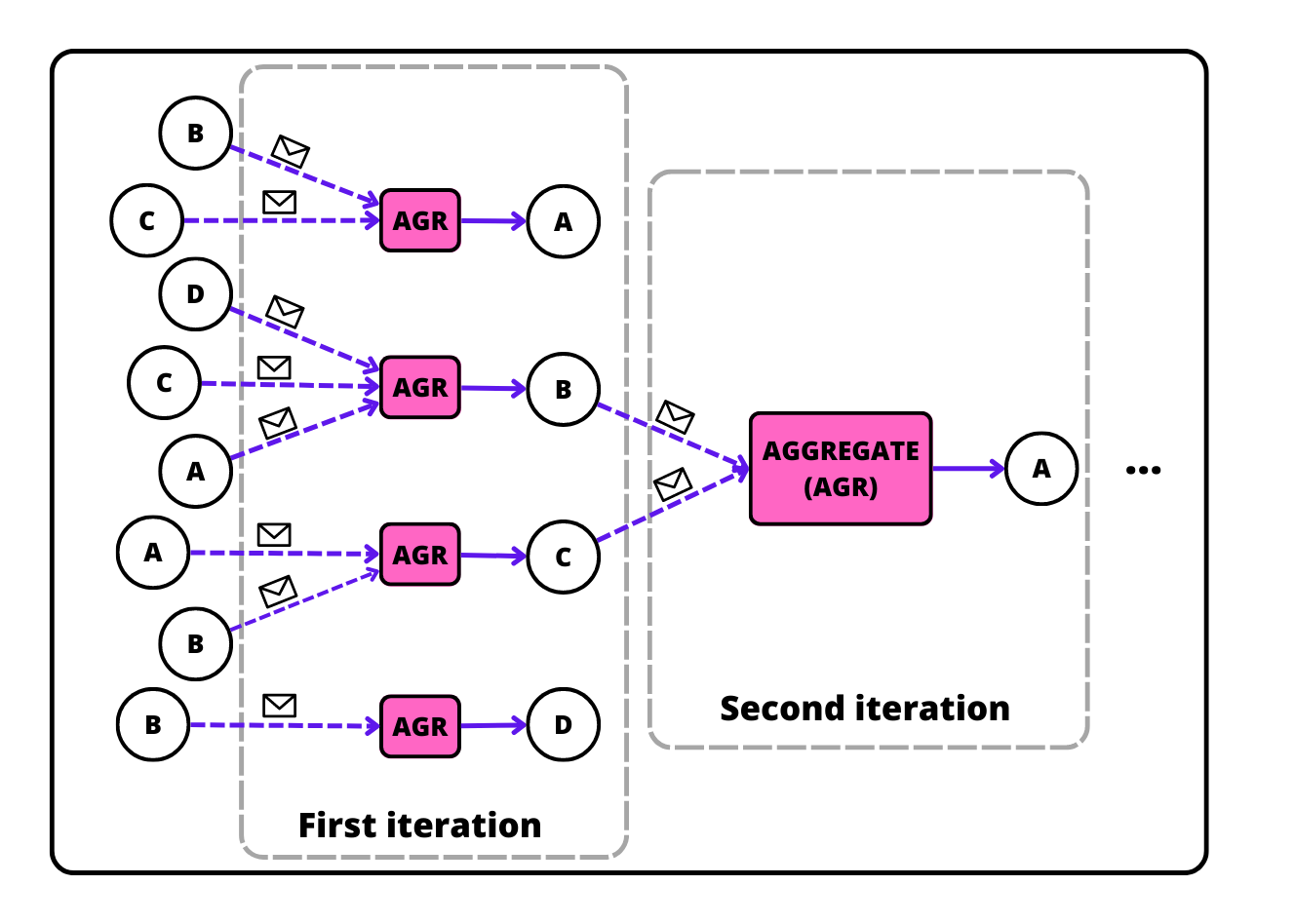}%
  }
  \caption{Aggregation of messages in a bidirectional graph: (a) input graph and (b) an example of GNN working mechanism (for the sake of simplification, the second iteration considers node `A'\ only.).}
  \label{fig:GNN_explanation}
\end{figure}

Graph Neural Networks bring the problem of learning patterns in a dataset to the graph domain. Formally, a graph ${\cal G} = ({\cal V}, {\cal E})$ is defined as a set of nodes $V{\cal }$ and a set of edges ${\cal E}$ between them, aka the adjacency relation. During the iteration process, each node (receiver) receives a set of aggregated messages from its neighbors, applying an aggregation function and an update function. It is worth noting that each node forwards information to its neighbors before its features are updated. In the next iteration, it forwards the new information (message) to its neighbors once more, as illustrated in Figure~\ref{fig:GNN_explanation}. 

For each iteration $k$, a hidden vector ${\textbf{h}_u}^{(k)}\in\mathbb{R}^n$ incorporates the features of node $u \in {\cal V}$, where $n$ stands for the number of input features. It is worth noting that the hidden vector ${\textbf{h}_u}^{(0)}$ encodes the features before training, i.e., at the initial stage. Firstly, a node-order invariant function is used to aggregate features from the neighborhood $\EuScript{N}(u)$ of node $u$. Secondly, the aggregated features are used to update the node information, described as follows:

\begin{equation}
\label{Eq1}
\textbf{h}_u^{k+1} = U^{k}~\left(\textbf{h}_u^{k},~A^{k}_u \left(\{\textbf{h}_v^{k}, ~\forall v \in \EuScript{N}(u)\}\right)\right),
\end{equation}
where $U^{(k)}$ and $A^{(k)}$ stand for the updating and aggregating functions, respectively. One can use distinct models for these functions, but this paper employs a formulation based on an attention mechanism, described further,

\subsection{Graph Attention Networks}
\label{ss.GAT}

Graph Attention Networks (GATs) are a strategy for improving the aggregation function. In this network, the message gives different priorities to the information from the neighborhood. The first application of this concept in a model was described by Veličković et al.~\cite{velivckovic2017graph} and crafted as follows:
\begin{equation}
\textbf{h}_u^{k+1} = \sigma\left(\sum_{v \in \EuScript{N}(u)}{ \alpha_{v\rightarrow u}^{k}\textbf{W}^{k}\textbf{h}_v^{k}}\right),
\end{equation}
where $\textbf{W}\in\mathbb{R}^{ n'\times n}$ is a trainable parameter known as the weight matrix, $n'$ and $\sigma$ stand for the number of output features and the sigmoid function, respectively. In addition, $\alpha_{v\rightarrow u}\in\mathbb{R}$ indicates the attention given from $v$ to the node $u$, i.e., the degree of influence $v$ has on updating the features of node $u$. A higher value of $\alpha_{v\rightarrow u}$ implies a stronger impact of $v$ on the feature update process of $u$. Formally, its definition is represented as follows:
\begin{equation}
\label{e.alpha}
    \alpha_{v\rightarrow u}^{k} = \frac{exp\left(\lambda\left(\left[\textbf{a}_u^{k}\right]^T\left[\textbf{W}^{k}\textbf{h}_u^{k} \mathbin\Vert \textbf{W}^{k}\textbf{h}_v^{k}\right]\right) \right)}{\sum_{v' \in \EuScript{N}(u)}{ exp\left(\lambda\left(\left[\textbf{a}_u^{k}\right]^T\left[\textbf{W}^{k}\textbf{h}_u^{k} \mathbin\Vert \textbf{W}^{k}\textbf{h}_{v'}^{k}\right]\right)\right)}},
\end{equation}
where $\textbf{a}_u\in\mathbb{R}^{2*n'}$ defines a trainable parameter known as the attention vector. The symbol $\mathbin\Vert$ denotes the concatenation operator, and $\lambda$ represents the LeakyReLU non-linearity function (with negative input slope $\beta = 0.2$). 

In addition, this particular GNN has proven to be more effective in accurately identifying the healthy state of patients by analyzing the facial landmarks extracted from their expressions during task performance.

\section{Experimental Methodology}
\label{s.methodology}

\subsection{Dataset}
\label{ss.dataset}

Established by Bandini et al. \cite{bandini2020new}, Toronto NeuroFace is the first public dataset with videos of oro-facial gestures performed by individuals with oro-facial impairments, including post-stroke (PS), ALS, and healthy control (HC). The dataset comprises 261 colored (RGB) videos of thirty-six participants: 11 patients with ALS, 14 patients with PS, and 11 HC. This work focuses on distinguishing ALS from healthy individuals, for we are primarily interested in the former. Therefore, we concentrated on a subset containing ALS and HC groups only. Each video captures a participant performing one of the subtasks from a set of speech and non-speech tasks commonly used during the clinical oro-facial examination. After manually segmenting the videos, we divided the dataset into 921 videos of repetitions. Table~\ref{t.subtask_repetition} presents the distribution of the number of repetitions for each subtask used in the experiments.

\begin{table}[ht]
\caption{Number of repetitions for each subtask.}
\begin{tabular}{|c|l|c|c|}
\hline
\textbf{Subtask}                                 & \multicolumn{1}{c|}{\textbf{Description}}                                                                                                 & \textbf{ALS}          & \textbf{HC}           \\ \hline
\cellcolor[HTML]{FFFFFF}                         &                                                                                                                                           &                       &                       \\
\multirow{-2}{*}{\cellcolor[HTML]{FFFFFF}SPREAD} & \multirow{-2}{*}{Pretending to smile with tight lips}                                                                                     & \multirow{-2}{*}{55}  & \multirow{-2}{*}{59}  \\ \hline
                                                 &                                                                                                                                           &                       &                       \\
\multirow{-2}{*}{KISS}                           & \multirow{-2}{*}{Pretend to kiss a baby}                                                                                                  & \multirow{-2}{*}{59}  & \multirow{-2}{*}{57}  \\ \hline
\cellcolor[HTML]{FFFFFF}                         &                                                                                                                                           &                       &                       \\
\multirow{-2}{*}{\cellcolor[HTML]{FFFFFF}OPEN}   & \multirow{-2}{*}{Maximum opening of the jaw}                                                                                              & \multirow{-2}{*}{54}  & \multirow{-2}{*}{55}  \\ \hline
                                                 &                                                                                                                                           &                       &                       \\
\multirow{-2}{*}{BLOW}                           & \multirow{-2}{*}{Pretend to blow a candle}                                                                                                & \multirow{-2}{*}{31}  & \multirow{-2}{*}{39}  \\ \hline
                                                 &                                                                                                                                           &                       &                       \\
\multirow{-2}{*}{BBP}                            & \multirow{-2}{*}{\begin{tabular}[c]{@{}l@{}}Repetitions of the sentence \\ “Buy Bobby a Puppy”\end{tabular}}                              & \multirow{-2}{*}{95}  & \multirow{-2}{*}{111} \\ \hline
                                                 &                                                                                                                                           &                       &                       \\
\multirow{-2}{*}{PA}                             & \multirow{-2}{*}{\begin{tabular}[c]{@{}l@{}}Repetitions of the syllables /pa/ \\ as fast as possible in a single breath\end{tabular}}     & \multirow{-2}{*}{100} & \multirow{-2}{*}{110} \\ \hline
                                                 &                                                                                                                                           &                       &                       \\
\multirow{-2}{*}{PATAKA}                         & \multirow{-2}{*}{\begin{tabular}[c]{@{}l@{}}Repetitions of the syllables /pataka/ \\ as fast as possible in a single breath\end{tabular}} & \multirow{-2}{*}{88}  & \multirow{-2}{*}{108} \\ \hline
\end{tabular}
\label{t.subtask_repetition}
\end{table}

\subsection{Pre-processing}
\label{ss.pre}

\begin{figure}[!htp]
	\centering
	\includegraphics[trim=0 5cm 0 5cm,clip, width=1\columnwidth]{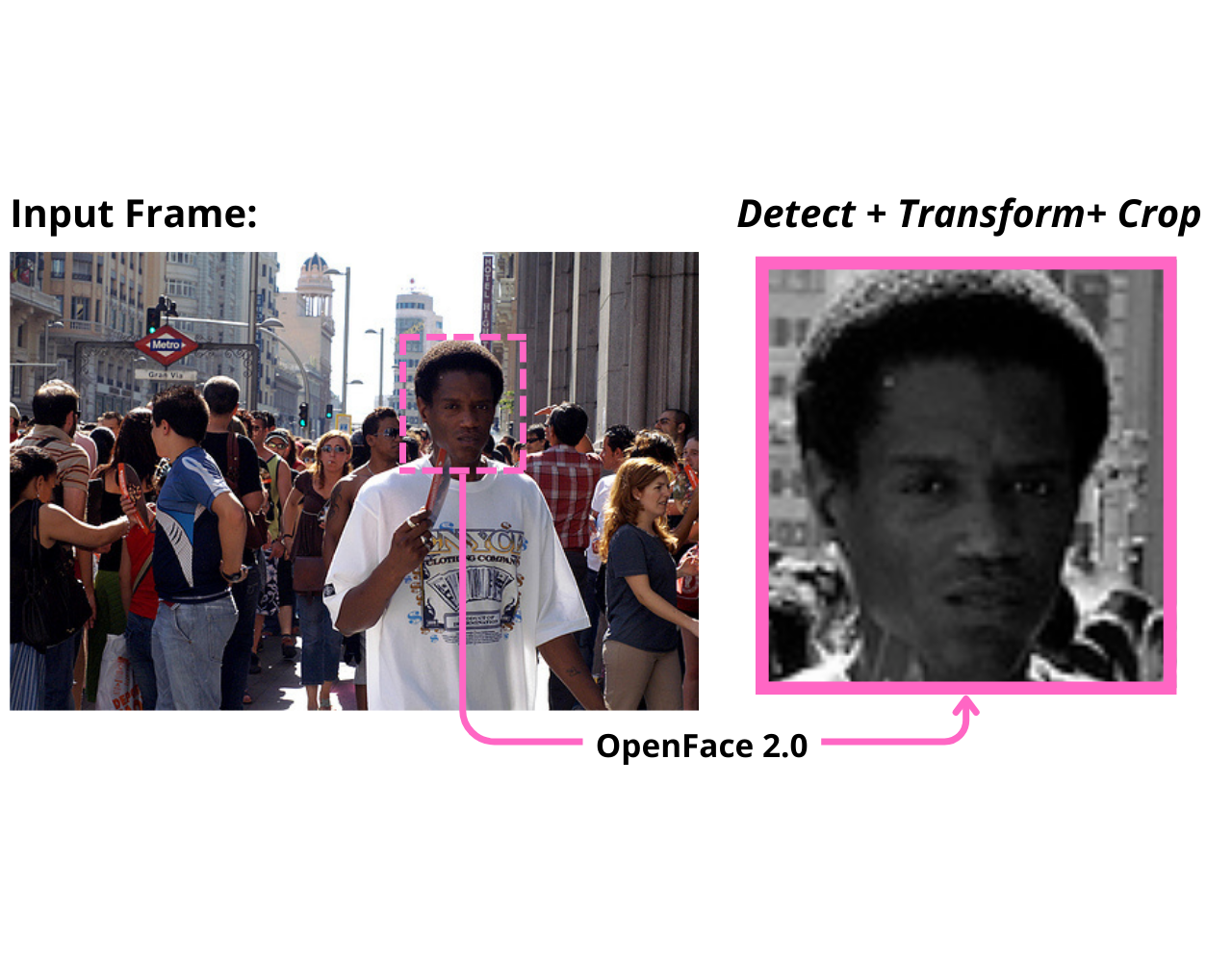}
	\caption{Illustration of OpenFace 2.0 for: (i) main face detection, (ii) transformation based on head pose estimation, and (iii) face cropping on an image from the Flickr30k dataset~\cite{young2014image}.}
	\label{f.pre-processing}
\end{figure}

To eliminate visual elements outside the subject's face and ensure consistency in the dataset, we use OpenFace 2.0 tool \cite{baltrusaitis2018openface} during the preprocessing stage. This tool first detects the main face, then perform a transformation based on head pose estimation and a crop operation on all frames. The resulting output ends up in $200\times200$ grayscale images centered on the facial region, as illustrated in Figure \ref{f.pre-processing}. 

\subsection{Feature extraction}
\label{ss.pre}

In this work, we used the Facial Alignment Network (FAN)~\cite{bulat2017far}, a deep learning model, to identify the frame-by-frame face geometric structure of each video in the dataset. As a state-of-the-art approach, FAN employs heatmap regression to accurately detect facial landmark points following the MULTI-PIE 2D 68-point format~\cite{gross2010multi}, enabling alignment in two and three dimensions. Since the dataset contains videos recorded with the frontal face position, we considered the alignment in two dimensions only.

\begin{figure}[!htp]
	\centering
	\includegraphics[trim=3.5cm 5cm 4cm 6cm,clip, width=1\columnwidth]{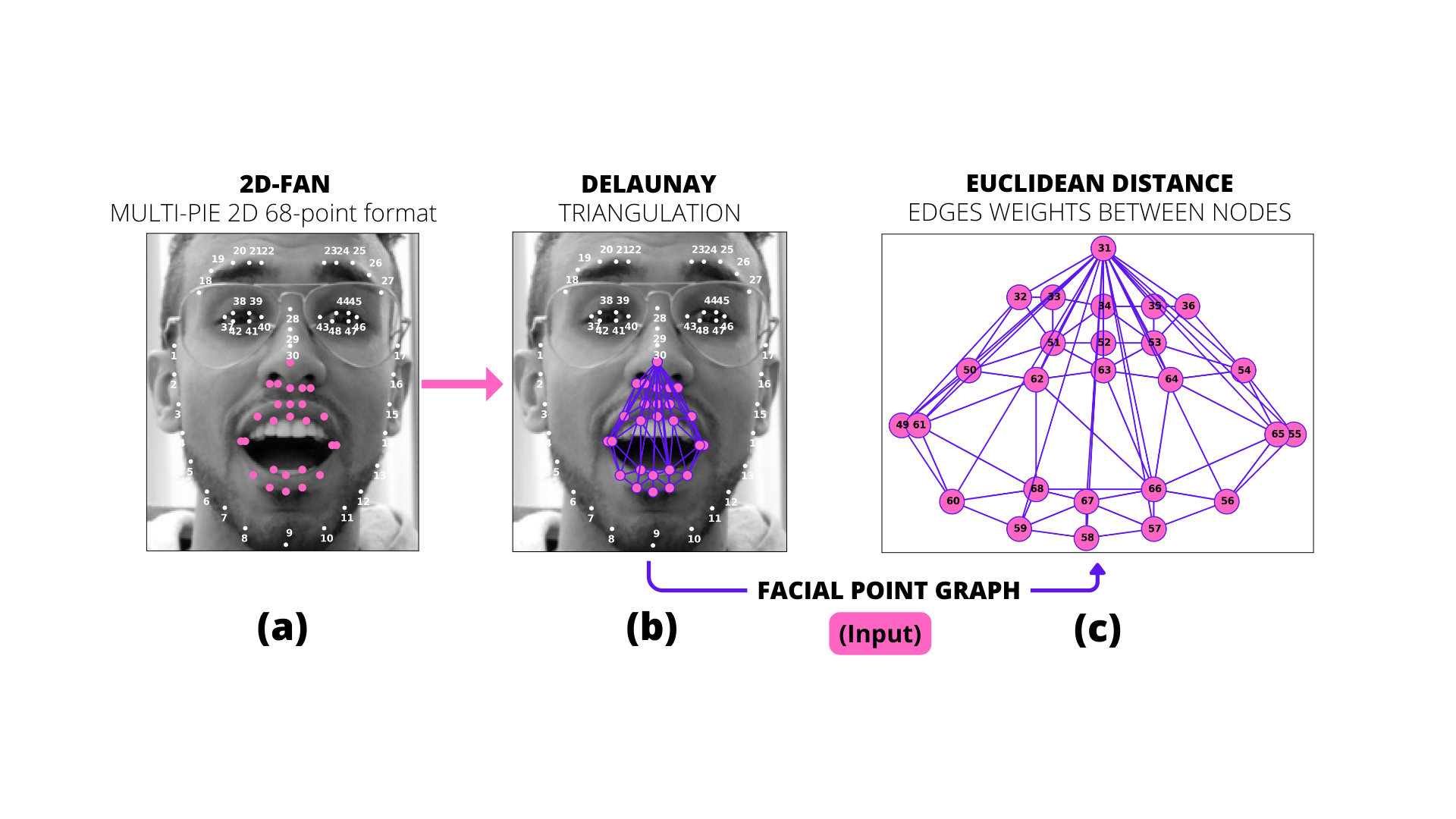}
	\caption{Representation of the feature extraction process.}
	\label{f.feature}
\end{figure}

Previous studies show that patients with ALS exhibit significant sensitivity in lip and jaw movements \cite{langmore1994physiologic, bandini2018kinematic}. Therefore, we selected 26 points from the landmarks extracted by FAN that represent such regions (Figure~\ref{f.feature}a). To establish connections between these landmark nodes, we employed the Delaunay triangulation ~\cite{delaunay1934sphere}, which involves creating a triangular mesh by connecting the specific landmarks (Figure~\ref{f.feature}b).

To enhance information communication among graph nodes during the learning process, we strategically use point 31 (According to the 68-point format)~\cite{gross2010multi}, which corresponds to the nose tip, as a central node. This key node serves as a hub, connecting all other nodes independently of the Delaunay triangulation calculation (Figure~\ref{f.feature}c). Lastly, as the final step of the feature extraction process, we set the edge's weight as the Euclidean distance between its corresponding nodes. 

\subsection{Classification and evaluation}

The classification performance was evaluated using a leave-one-subject-out cross-validation (LOSO-CV) approach, following the method proposed by Bandini et al.~\cite{bandini2018automatic}. Furthermore, to enhance the reliability of predictions in real-world scenarios and mitigate issues like overfitting or memorizing training data, we employ separate sets for training, validation, and testing in each interaction. Concerning the validation sets, we randomly select two subjects, one categorized as HC and the other as ALS, ensuring a balanced representation of both classes in this stage. 

The evaluation was conducted in two modes, i.e., repetition- and subject-based classification:

\subsubsection{Repetition classification}
\label{sss.rep}

\begin{figure}[!htp]
	\centering
	\includegraphics[trim=1cm 3cm 1cm 3cm,clip, width=1\columnwidth]{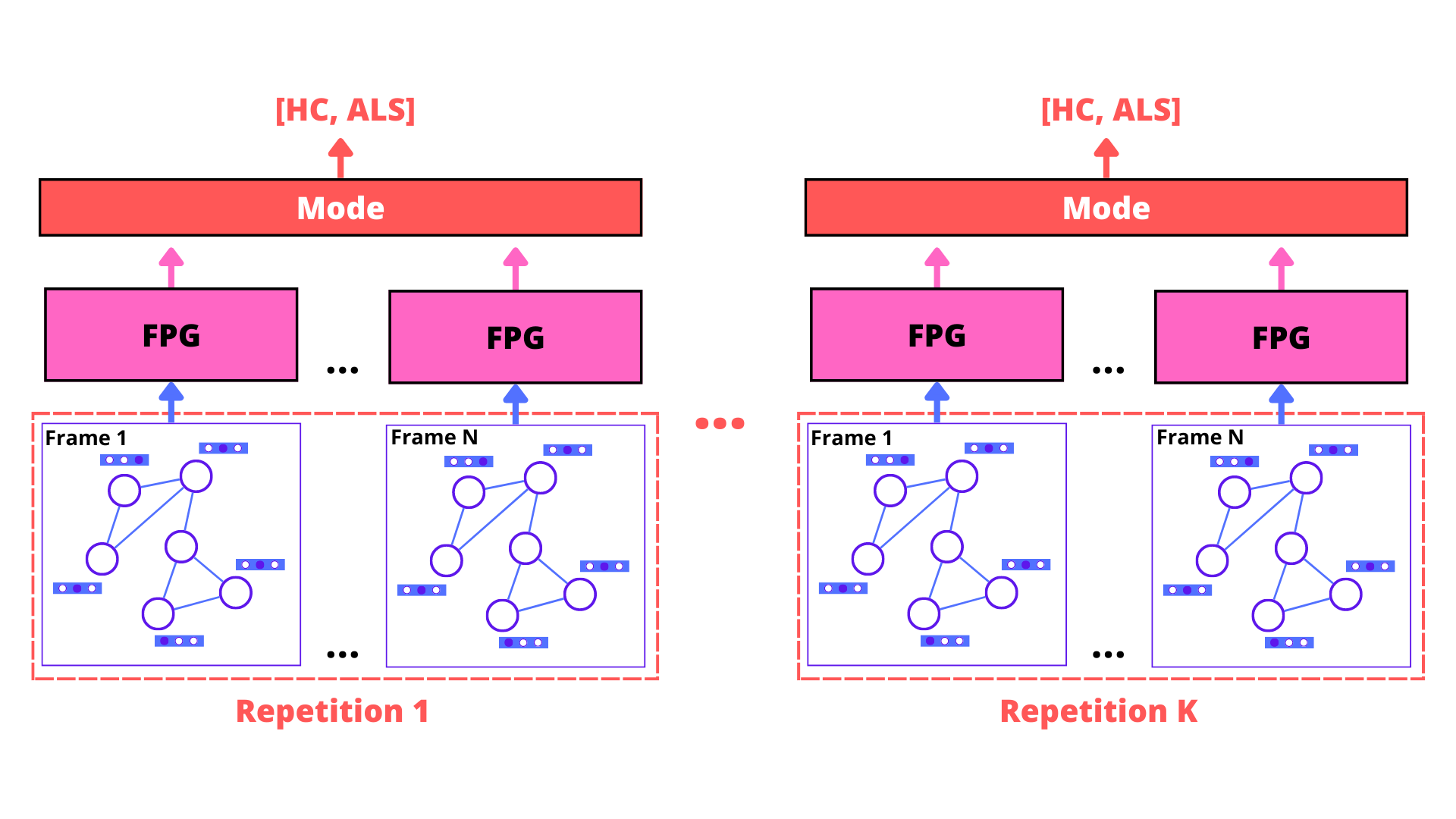}
	\caption{Overview of the repetition-based evaluation step.}
	\label{f.repetition}
\end{figure}

For each iteration of the LOSO-CV, the repetitions produced by one participant were treated as individual samples in the test set. At the same time, the remaining data was split into validation and training sets. This approach ensures that every participant, both HC and those with ALS, and their respective repetitions were considered in separate test sets. During this trial, individuals' speech and non-speech repetitions were classified as belonging to the HC or ALS group. Figure~\ref{f.repetition} illustrates the process mentioned above for a given individual\footnote{In this experiment, we count the hit/miss over each repetition (for each individual) to compose the final classification accuracy. Basically, we are labeling repetitions and not individuals.}.

\subsubsection{Subject classification}
\label{sss.sub}

\begin{figure}[!htb]
	\centering
	\includegraphics[trim= 1cm 3.5cm 1cm 4cm,clip, width=0.7\columnwidth]{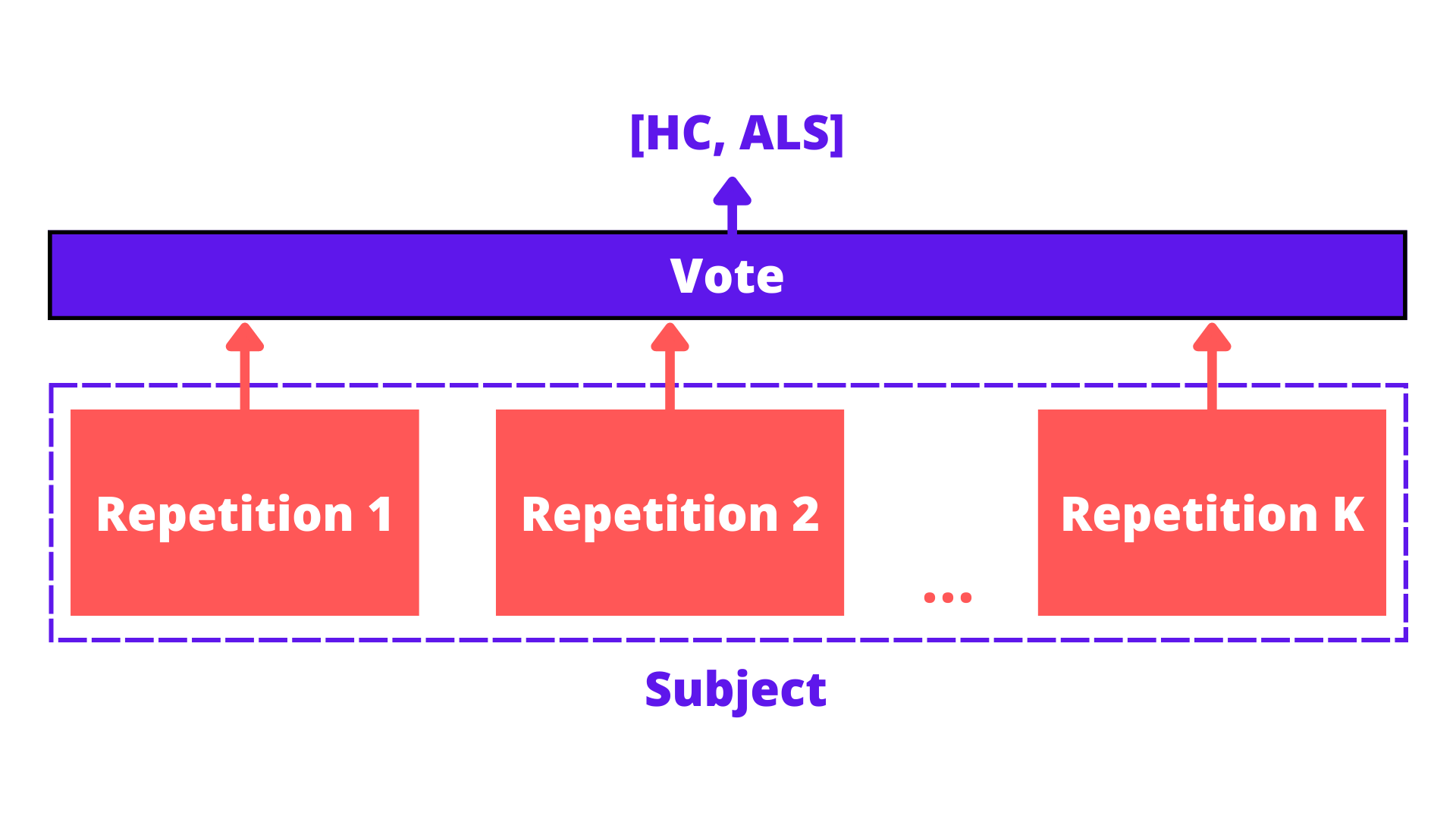}
	\caption{Overview of the subject-based evaluation step.}
	\label{f.subject}
\end{figure}

At each iteration of the LOSO-CV, each subject was treated as a test case and classified as either HC or ALS. The classification was determined through a majority vote among its predicted repetitions; in tie cases, the subject was considered HC to generate a more conservative prediction according to Bandini et al.~\cite{bandini2018automatic}. Figure~\ref{f.subject} depicts an overview of the subject classification process.

In both repetition- and subject-based classification, the validation set was used to prevent bias in the model's hyperparameters and to facilitate the implementation of the early stopping technique. The number of epochs for training was determined by monitoring the learning progress on the validation set\footnote{The maximum number of epochs is set to $100$, the batch-size comprises $64$ samples, the learning rates are set to $10^{-4}$ and $10^{-5}$ considering the GAT and linear layers, respectively. The number of hidden layers was set to $17$. These values were empirically chosen  based on the results over the validation set.}.

Considering Bandini et al.~\cite{bandini2018automatic} as the benchmark to our work, the experiments were also performed considering two other classification models for comparison purposes: SVM with linear and radial basis function (RBF) and Logistic Regression. Both models use 11 geometric and kinematic features extracted from speech and non-speech tasks. A grid search was used to find proper values for SVM parameters, i.e., the confidence value $C$ and the RBF kernel scale parameter $\gamma$.

\subsection{Proposed model}
\label{s.proposed}

Initially, the proposed model uses 15 equally spaced frames for each repetition performed by the patient. FPG receives a graph of twenty-six nodes representing the face landmarks, where each node encodes a feature vector with the $x$ and $y$ coordinates of its related landmark. In addition, each graph edge stores its length determined by the Euclidean distance between its two corresponding nodes.

\begin{figure*}
    \centering
	\includegraphics[width=0.85\paperwidth]{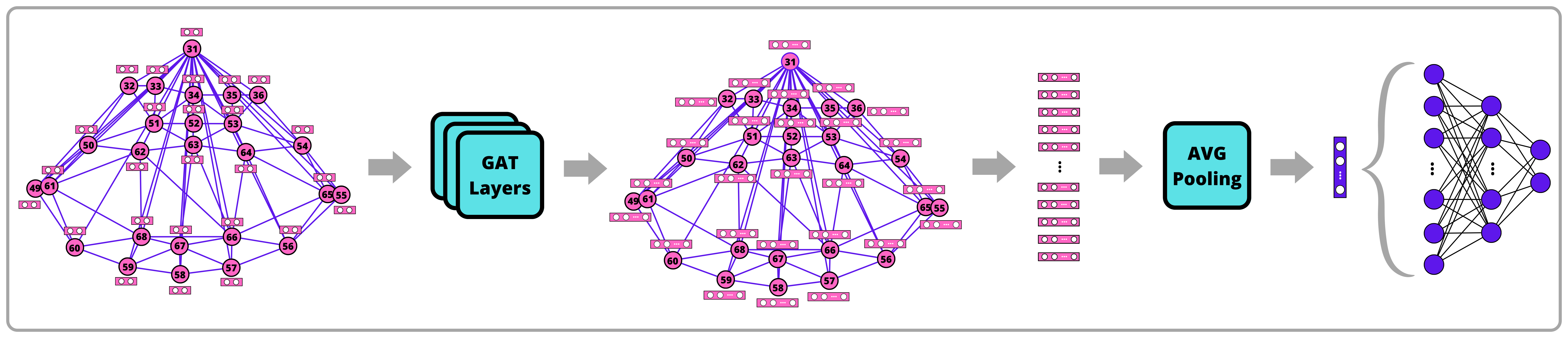}
	\caption{FPG model.}
	\label{f.proposed_model}
\end{figure*}

Further, each frame proceeds through six GAT and two linear layers. Before the information is forwarded to the linear layers, an average pooling is performed using the nodes, i.e., all information encoded in the graph is mapped into a single vector. Figure~\ref{f.proposed_model} illustrates such a process.

The result obtained after pooling goes through two linear layers, which generate the model's output. Nonetheless, the error is calculated based on the frame label, and the mode among frames represents the outcomes concerning the repetition experiment. In other words, classifying an individual's repetition is based on the majority consensus among the frames. Likewise, when classifying the subject, the majority mode derived from the classifications of each repetition determines whether the patient has ALS or not.





\section{Results}
\label{s.res}

The experimental results were obtained for each subtask separately. To provide a more in-depth evaluation of the proposed approach, we considered three evaluation measures: accuracy, sensitivity, and specificity. Table~\ref{t.results} presents the results for each subtask accordingly.

\begin{table}[!htb]
\caption{FPG results for each subtask in speech and non-speech data.}
\label{t.results}
\scalebox{0.91}{
\begin{tabular}{|cc|c|c|c|c|}
\hline
\multicolumn{2}{|c|}{TASK}                                                                                                        & Classification                              & Accuracy                                & Specificity                             & Sensitivity                             \\ \hline
\multicolumn{1}{|c|}{}                                                                        &                                   & \cellcolor[HTML]{FFFFFF}\textbf{Repetition} & \cellcolor[HTML]{FFFFFF}\textbf{80,7\%} & \cellcolor[HTML]{FFFFFF}\textbf{79,6\%} & \cellcolor[HTML]{FFFFFF}\textbf{81,8\%} \\ \cline{3-6} 
\multicolumn{1}{|c|}{}                                                                        & \multirow{-2}{*}{\textbf{SPREAD}} & \cellcolor[HTML]{F9B0E1}\textbf{Subject}    & \cellcolor[HTML]{F9B0E1}\textbf{81,8\%} & \cellcolor[HTML]{F9B0E1}\textbf{81,8\%} & \cellcolor[HTML]{F9B0E1}\textbf{81,8\%} \\ \cline{2-6} 
\multicolumn{1}{|c|}{}                                                                        &                                   & \cellcolor[HTML]{FFFFFF}Repetition          & \cellcolor[HTML]{FFFFFF}68,1\%          & \cellcolor[HTML]{FFFFFF}80,7\%          & \cellcolor[HTML]{FFFFFF}55,9\%          \\ \cline{3-6} 
\multicolumn{1}{|c|}{}                                                                        & \multirow{-2}{*}{KISS}            & \cellcolor[HTML]{F9B0E1}Subject             & \cellcolor[HTML]{F9B0E1}68,1\%          & \cellcolor[HTML]{F9B0E1}81,8\%          & \cellcolor[HTML]{F9B0E1}54,5\%          \\ \cline{2-6} 
\multicolumn{1}{|c|}{}                                                                        &                                   & \cellcolor[HTML]{FFFFFF}Repetition          & \cellcolor[HTML]{FFFFFF}77,0\%          & \cellcolor[HTML]{FFFFFF}78,1\%          & \cellcolor[HTML]{FFFFFF}75,9\%          \\ \cline{3-6} 
\multicolumn{1}{|c|}{}                                                                        & \multirow{-2}{*}{OPEN}            & \cellcolor[HTML]{F9B0E1}Subject             & \cellcolor[HTML]{F9B0E1}81,8\%          & \cellcolor[HTML]{F9B0E1}81,8\%          & \cellcolor[HTML]{F9B0E1}81,8\%          \\ \cline{2-6} 
\multicolumn{1}{|c|}{}                                                                        &                                   & \cellcolor[HTML]{FFFFFF}Repetition          & \cellcolor[HTML]{FFFFFF}37,1\%          & \cellcolor[HTML]{FFFFFF}51,2\%          & \cellcolor[HTML]{FFFFFF}19,3\%          \\ \cline{3-6} 
\multicolumn{1}{|c|}{\multirow{-8}{*}{\begin{tabular}[c]{@{}c@{}}Non-\\ speech\end{tabular}}} & \multirow{-2}{*}{BLOW}            & \cellcolor[HTML]{F9B0E1}Subject             & \cellcolor[HTML]{F9B0E1}38,4\%          & \cellcolor[HTML]{F9B0E1}57,1\%          & \cellcolor[HTML]{F9B0E1}16,6\%          \\ \hline
\multicolumn{1}{|c|}{}                                                                        &                                   & \cellcolor[HTML]{FFFFFF}Repetition          & \cellcolor[HTML]{FFFFFF}49,0\%          & \cellcolor[HTML]{FFFFFF}63,0\%          & \cellcolor[HTML]{FFFFFF}32,6\%          \\ \cline{3-6} 
\multicolumn{1}{|c|}{}                                                                        & \multirow{-2}{*}{BBP}             & \cellcolor[HTML]{F9B0E1}Subject             & \cellcolor[HTML]{F9B0E1}50,0\%          & \cellcolor[HTML]{F9B0E1}63,6\%          & \cellcolor[HTML]{F9B0E1}33,3\%          \\ \cline{2-6} 
\multicolumn{1}{|c|}{}                                                                        &                                   & \cellcolor[HTML]{FFFFFF}Repetition          & \cellcolor[HTML]{FFFFFF}64,2\%          & \cellcolor[HTML]{FFFFFF}64,5\%          & \cellcolor[HTML]{FFFFFF}64,0\%          \\ \cline{3-6} 
\multicolumn{1}{|c|}{}                                                                        & \multirow{-2}{*}{PA}              & \cellcolor[HTML]{F9B0E1}Subject             & \cellcolor[HTML]{F9B0E1}57,1\%          & \cellcolor[HTML]{F9B0E1}54,5\%          & \cellcolor[HTML]{F9B0E1}60,0\%          \\ \cline{2-6} 
\multicolumn{1}{|c|}{}                                                                        &                                   & \cellcolor[HTML]{FFFFFF}Repetition          & \cellcolor[HTML]{FFFFFF}67,3\%          & \cellcolor[HTML]{FFFFFF}65,7\%          & \cellcolor[HTML]{FFFFFF}69,3\%          \\ \cline{3-6} 
\multicolumn{1}{|c|}{\multirow{-6}{*}{Speech}}                                                & \multirow{-2}{*}{PATAKA}          & \cellcolor[HTML]{F9B0E1}Subject             & \cellcolor[HTML]{F9B0E1}66,6\%          & \cellcolor[HTML]{F9B0E1}63,6\%          & \cellcolor[HTML]{F9B0E1}70,0\%          \\ \hline
\end{tabular}}
\end{table}

According to previous studies, the SPREAD subtask also appears to be the most discriminative one, with an accuracy of 80.7\% during repetition-based classification and 81.8\% concerning the subject-based classification in our model approach. As described in Section~\ref{s.methodology}, the experiments were conducted by first splitting the dataset into training and test folds. The former was partitioned into a smaller training set to generate a validation fold, whose size was limited to the data available for training. 

We anticipate that the results obtained using SVM and Logistic Regression may differ significantly from the findings presented by Bandini et al.~\cite{bandini2018automatic}, for they employed a slightly different approach. Although Toronto Neuroface contains the same speech and non-speech tasks as those in the study conducted by Bandini et al.~\cite{bandini2018automatic}, our approach has several differences. Firstly, the participants in the dataset we had access to were not the same, and there were variations in terms of quantity. Moreover, we manually cropped the frames containing repetitions, for only the entire video was made available. Additionally, we did not have access to the videos containing samples from the REST subtask, which was used for normalization in both SVM and Regression models. Furthermore, our videos only included color information and did not incorporate three-dimensional depth features.

\begin{figure*}[!htb]
    \centering
	\includegraphics[width=0.85\paperwidth]{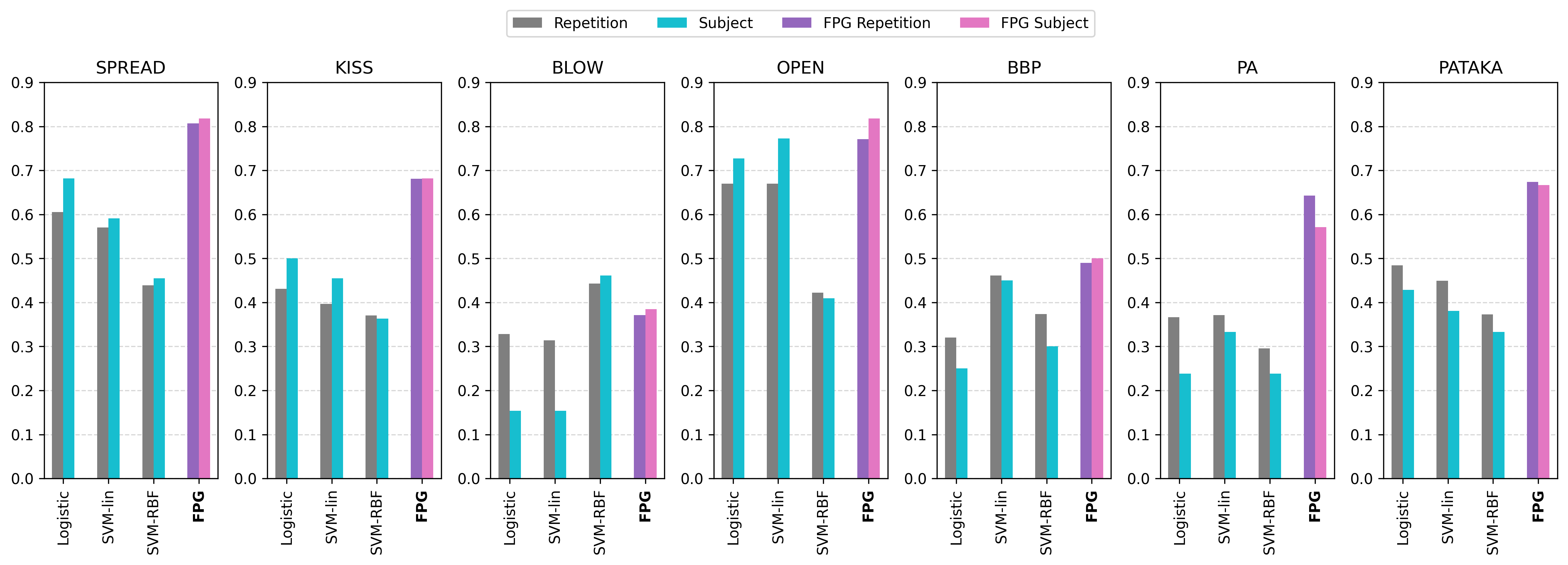}
	\caption{Comparison between FPG and baselines.}
	\label{f.non_speech}
\end{figure*}

Figure~\ref{f.non_speech} compares FPG against the baselines inspired in Bandini et al.~\cite{bandini2018automatic} work. One can observe that our model consistently outperforms others in the majority of tasks, e.g., SPREAD, KISS, PA, and PATAKA. However, SVM-RBF stands out as the top-performing model in the BLOW subtask. However, SVM-RBF stands out as the top-performing model in the BLOW subtask, which was the most challenging, as also observed by Bandini et al.~\cite{bandini2018automatic}.

\section{Discussion and Conclusion}
\label{s.discussion_conclusion}

To the best of our knowledge, the current study is the first to evaluate Graph Neural Networks for ALS identification based on facial expression. As the main finding, we showed state-of-the-art results in all subtasks of the Toronto Neuroface dataset but one.


The two highest accuracies are observed in SPREAD and OPEN subtasks, achieving results above 80\%. We can observe similar values for the specificity and sensitivity in both subtasks, showing the model's robustness in distinguishing ALS patients from healthy ones.


We attribute the high accuracy in the SPREAD task to the pure lip movement not involving the jaw muscles~\cite{bandini2018automatic}, allowing the detection of the loss of lip muscle extension exhibited by bulbar ALS patients. Additionally, as shown in previous studies, the jaw muscles decline in bulbar ALS patients~\cite{bandini2018automatic}. Consequently, the extension of this movement was distinguished with high accuracy by the model during the OPEN task. OPEN considers the greatest extent of jaw muscle movement among all other tasks, justifying the model's accuracy.


The exchange of information among the graph nodes during the learning iterations allowed for better differentiation of facial points between individuals with ALS or HC. It is also noteworthy that, except for PA and PATAKA tasks, the model showed inferior or equal results in repetition classification compared to subject-base classification, indicating that most repetitions were correctly classified, as the mode of labeled repetitions ended in the correct classification of the subject.


One of the major limitations and challenges in training deep learning models is the limited number of videos available in the dataset. Deep models typically require a substantial amount of data to learn effectively. However, FPG showcased exceptional performance despite being a deep approach. Remarkably, it achieved high accuracy without data augmentation during training.

Such outcomes highlight the effectiveness of GNN models, showcasing their inherent structural characteristics and information propagation capabilities. GNNs demonstrate their ability to capture complex patterns and relationships within the data, even when dealing with a limited dataset, underscoring GNNs as a powerful approach in this particular domain.


This study did not consider the order of repetitions. Therefore, exploring temporal information in the Facial Point Graph as a future work would be interesting, particularly the changes observed in facial movements in the presence of neurodegenerative diseases. We shall further investigate the impact of fatigue found in ALS patients during speech tasks like 'PA' and 'PATAKA' and potentially improve the model's performance in capturing these variations.



Despite the limitations and complexity of the problem, the proposed approach achieved significant results when compared to similar works, introducing the Facial Point Graph for ALS diagnosis. In addition, the results were achieved without handcrafted features and with a lightweight model, enabling the development of affordable systems capable of supporting clinicians in automatic ALS diagnosis.

\section*{Acknowledgment}
The authors would like to thank FAPESP (Fundação de Amparo à Pesquisa do Estado de São Paulo), process number \#2022/13156-8 and \#2022/16090-8.



\bibliographystyle{IEEEtran}

%
%


\end{document}